\documentclass[sigconf]{acmart}
\acmConference[ICSE 2024]{46th International Conference on Software Engineering}{April 2024}{Lisbon, Portugal}
\usepackage{hyperref}       
\usepackage{url}            
\usepackage{booktabs}       
\usepackage{amsfonts}       
\usepackage{nicefrac}       
\usepackage{microtype}      
\usepackage{xcolor}         
\usepackage{mathtools} 
\usepackage{booktabs} 
\usepackage{xspace}
\usepackage{graphicx}
\usepackage{subfigure}
\usepackage{cleveref}
\usepackage{enumitem}
\usepackage[linesnumbered,ruled,vlined]{algorithm2e}

\SetCommentSty{mycommfont}
\usepackage{tabularray}
\usepackage{xcolor}
\usepackage{wrapfig}
\usepackage{epigraph}
\usepackage{multirow}
\usepackage{colortbl}
\usepackage{tikz}

\usepackage{hyperref}
\usepackage{url}
\usepackage{diagbox}
\usepackage[many]{tcolorbox}

\usepackage{tabularray}

\usepackage{xspace}
\makeatletter
\DeclareRobustCommand\onedot{\futurelet\@let@token\@onedot}
\def\@onedot{\ifx\@let@token.\else.\null\fi\xspace}

\def\eg{\emph{e.g}\onedot} \def\Eg{\emph{E.g}\onedot}
\def\ie{\emph{i.e}\onedot} 
 
\def\etc{\emph{etc}\onedot} 
 
\def\etal{\emph{et al}\onedot}
\makeatother

\AtBeginDocument{%
  }

\setcopyright{acmcopyright}
\copyrightyear{2018}
\acmYear{2018}
\acmDOI{XXXXXXX.XXXXXXX}

\acmJournal{JDS}
\acmVolume{37}
\acmNumber{4}
\acmArticle{111}
\acmMonth{8}

\begin{document}

\title{LLMs for Relational Reasoning: How Far are We?}

\author{Zhiming Li}
\email{zhiming001@e.ntu.edu.sg}
\affiliation{%
  \institution{Continental-NTU Corporate Lab, Nanyang Technological University}
  \country{Singapore}
}

\author{Yushi Cao}\authornote{Corresponding author}
\email{yushi002@e.ntu.edu.sg}
\affiliation{%
  \institution{Continental-NTU Corporate Lab, Nanyang Technological University}
  \country{Singapore}}

\author{Xiufeng Xu}
\email{xiufeng001@e.ntu.edu.sg}
\affiliation{%
  \institution{Nanyang Technological University}
  \country{Singapore}}

\author{Junzhe Jiang}
\email{junzhe.jiang@connect.polyu.hk}
\affiliation{%
  \institution{Hong Kong Polytechnic University}
  \country{Hong Kong}}

\author{Xu Liu}
\email{liuxu@comp.nus.edu.sg}
\affiliation{%
  \institution{National University of Singapore}
  \country{Singapore}}

\author{Yon Shin Teo}
\email{yon.shin.teo@continental-corporation.com}
\affiliation{%
  \institution{Continental Automotive Singapore Pte. Ltd.}
  \country{Singapore}
}

\author{Shang-wei Lin}
\email{shang-wei.lin@ntu.edu.sg}
\affiliation{%
  \institution{Continental-NTU Corporate Lab, Nanyang Technological University}
  \country{Singapore}}

\author{Yang Liu}
\email{yangliu@ntu.edu.sg}
\affiliation{%
  \institution{Continental-NTU Corporate Lab, Nanyang Technological University}
  \country{Singapore}}
\renewcommand{\shortauthors}{Li et al.}
\acmArticleType{Review}
\acmCodeLink{https://github.com/borisveytsman/acmart}
\acmDataLink{htps://zenodo.org/link}

\def \VersionWithComments {}
\newcommand{\marginX}{\marginnote{\huge{\quad\quad\textbf{!}\quad\quad}}}
\newcommand{\lzm}[1]{\mbox{}{\color{blue}\marginX{}\textbf{[lzm}: #1]}}
\newcommand{\jjz}[1]{\textcolor{red}{jjz:#1}}
\newcommand{\xiu}[1]{\textcolor{purple}{Xiu: #1}\xspace}
\newcommand{\wbz}[1]{\textcolor{blue}{wbz: #1}\xspace}
\newcommand{\cys}[1]{\textcolor{green}{cys:#1}}
\newcommand{\lsw}[1]{\mbox{}{\color{orange}\marginX{}\textbf{[Shang-Wei}: #1]}}
\newcommand{\tys}[1]{\mbox{}{\color{pink}\textbf{[YonShin}: #1]}}
\newcommand{\instructions}[1]{{\color{red}\marginX{}\textbf{[Instructions: ``#1'']}}}
\newcommand{\reviewer}[2]{\mbox{}{\color{red}\marginX{}\textbf{[Reviewer #1}: ``#2'']}}
\newcommand{\todo}[1]{\mbox{}{\color{blue}{\marginX{}\textbf{TODO}\ifx#1\\\else:\ \fi #1}}}


\begin{abstract}

Large language models (LLMs) have revolutionized many areas (\eg~natural language processing, software engineering, \etc) by achieving state-of-the-art performance on extensive downstream tasks. Aiming to achieve robust and general artificial intelligence, there has been a surge of interest in investigating the reasoning ability of the LLMs. Whereas the textual and numerical reasoning benchmarks adopted by previous works are rather shallow and simple, it is hard to conclude that the LLMs possess strong reasoning ability by merely achieving positive results on these benchmarks. 
Recent efforts have demonstrated that the LLMs are poor at solving sequential
decision-making problems that require common-sense planning by evaluating their performance on the reinforcement learning benchmarks. 
In this work, we conduct an in-depth assessment of several state-of-the-art LLMs' reasoning ability based on the inductive logic programming (ILP) benchmark, which is broadly recognized as a representative and challenging measurement for evaluating logic program induction/synthesis systems as it requires inducing strict \emph{cause-effect} logic to achieve robust deduction on independent and identically distributed (IID) and out-of-distribution (OOD) test samples.
Our evaluations illustrate that compared with the neural program induction systems which are much smaller in model size, the state-of-the-art LLMs are much poorer in terms of reasoning ability by achieving much lower performance and generalization using either natural language prompting or truth-value matrix prompting\footnote{The implementation is available at: \url{https://sites.google.com/view/llm-rr}}.
\end{abstract}

\maketitle

\section{Introduction}
Large language models (LLMs) have achieved great breakthroughs in various domains such as natural language processing~\cite{gpt3,touvron2023llama}, software engineering~\cite{huang2023empirical,deng2023pentestgpt}, finance~\cite{wu2023bloomberggpt,yang2023fingpt}, \etc. There has been a recent increase in interest in exploring the reasoning ability of LLMs, which is regarded as a crucial ability of Artificial General Intelligence (AGI)~\cite{kojima2022large,wei2022chain,li2021implicit}. 
Prevalent reasoning ability evaluation benchmarks adopted by previous literature include arithmetic~\cite{cobbe2021arithmetic,patel2021arithmetic}, symbolic reasoning~\cite{wei2022chain,srivastava2022beyond}, commonsense~\cite{talmor2018commonsenseqa,sakaguchi2021winogrande}~\etc.
And with the recent advancement of the in-context few-shot (zero-shot) \emph{prompting} techniques~\cite{wei2022chain,gpt3,lester2021power}, LLMs manage to achieve state-of-the-art few-shot (zero-shot) learning performance on these benchmarks without training. Concisely, in-context few-shot (zero-shot) \emph{prompting} refers to the techniques that provide input to the language model to boost the performance of specific tasks. The input can be a few examples (few-shot)~\cite{brown2020language, wei2022chain} or merely instructions that describe the task (zero-shot)~\cite{wei2021finetuned}.
Despite the stunning achievement of the LLMs on these reasoning benchmarks. There are recent debate that these benchmarks are relatively simple in terms of task-solving logic and only require shallow reasoning to accomplish~\cite{valmeekam2023planbench}. Thus it is insufficient to support the claims about LLMs’ reasoning ability.
Specifically, Valmeekam \etal~\cite{valmeekam2023planbench} conducted evaluations of the LLMs' inherent emergent planning on a reinforcement learning benchmark: \emph{blocksworld}. The \emph{blocksworld} is a popular benchmark for evaluating reinforcement learning baselines in terms of sequential decision-making ability. Concretely, given a set of blocks, the goal of this task is to arrange the blocks in a particular order. While this task is simple enough for humans to solve, it is found that even some of the current state-of-the-art LLMs present poor performance on them.

Going beyond the evaluation of LLMs' sequential decision-making abilities on reinforcement learning benchmarks, we emphasize that the relational reasoning ability is another crucial ability to focus on. It is considered a crucial component of intelligence that directly correlates with the capacity to think logically and solve problems in novel situations~\cite{crone2009neurocognitive,halford1998processing}. Specifically, relational reasoning ability is the ability to reason about relationships between objects. 

The inductive logic programming (ILP) benchmarks~\cite{evans2018learning,dongneural,zimmer2023differentiable} are broadly used for the evaluation of program induction/synthesis systems' relational reasoning ability. In specific, ILP is a task that aims to automatically induce a logic program given
some positive examples and negative examples as specifications~\cite{evans2018learning}. In this work, to achieve comprehensive evaluations of the LLMs' relational reasoning ability, we develop a universal evaluation pipeline that enables detailed evaluation of both the state-of-the-art LLMs and the neural program induction baseline methods which are dedicatedly designed for relational reasoning. To the best of our knowledge, we are the first to conduct a detailed analysis of the LLMs' relational reasoning ability as well as their comparison with the neural program induction models. We have the following key findings:

\begin{itemize}
[topsep=2pt,itemsep=2pt,partopsep=0ex,parsep=0ex,leftmargin=*]
  \item We implement a universal evaluation pipeline for relational reasoning ability assessment, which is general for the evaluation of the state-of-the-art LLMs and the neural program induction models.
  \item We conduct comprehensive evaluations of LLMs' relational reasoning ability, which is the first of its kind.
  \item We unveil that the current state-of-the-art LLMs' relational reasoning ability is still far from perfect and is inferior compared to the neural program induction models which are much smaller in size.
\end{itemize}


\begin{figure*}[!ht] 
\centering 
\includegraphics[width=0.7\textwidth]{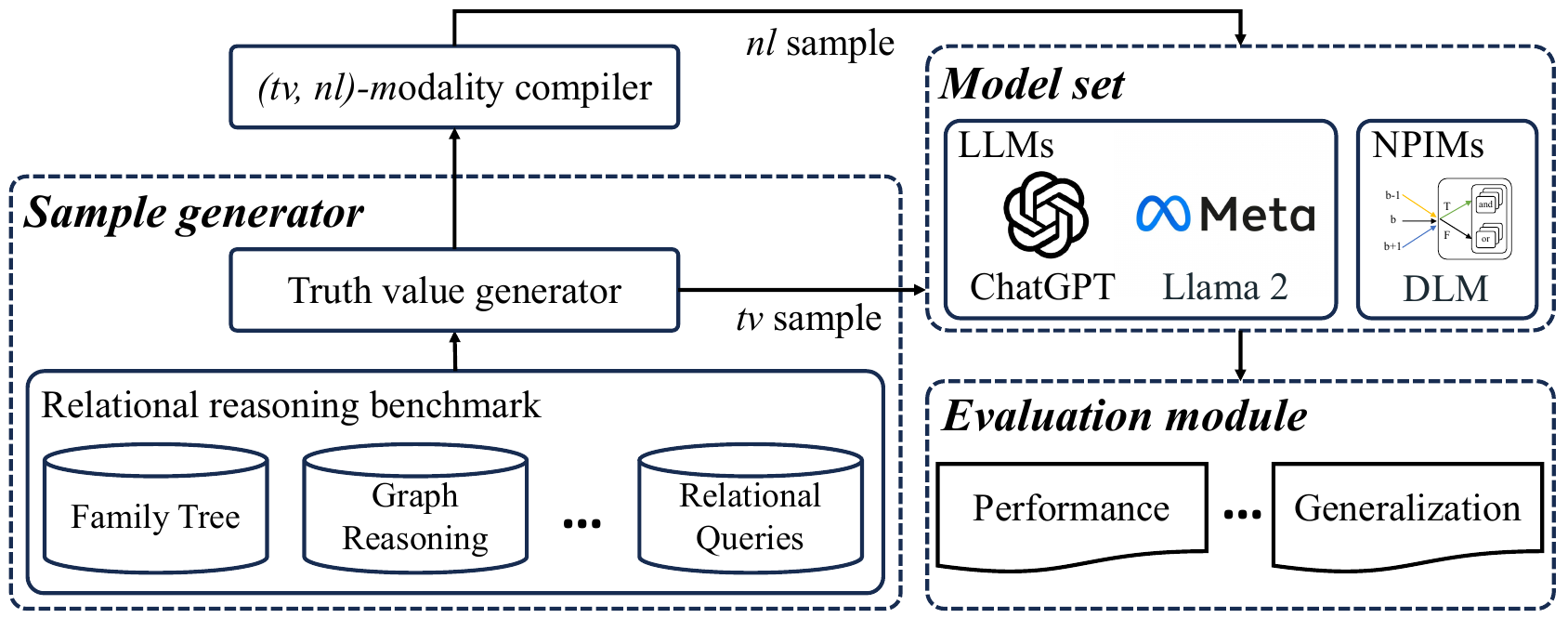} 
\caption{Overview of the evaluation pipeline.} 
\label{fig:overview} 
\vspace{-1em}
\end{figure*}

\section{Preliminaries}

\subsection{Inductive Logic Programming}
Program induction \& program synthesis~\cite{devlin2017robustfill,balog2016deepcoder,dongneural} are tasks that aim to induce a program that satisfies a given specification. Program induction requires the model to induce the program implicitly and conduct inference directly by the model itself without executing an explicit program, while program synthesis requires the model to generate an explicit program and execute it for inference. Inductive logic programming (ILP)~\cite{muggleton1991inductive,koller2007introduction,getoor2001learning} is a sub-field in which specification comes in the form of input-output examples. It requires the model to express the program with first-order logic rules.
ILP has long been considered a task that is demanding in logical (relational) reasoning ability to accomplish. Specifically, given a set of background predicates $\mathbf{B}$ (also called \emph{premise}), a set of positive examples $\rho$, and a set of negative samples $\eta$, the goal of a program induction/synthesis model is to derive a logic program $C$ that satisfies the following two conditions: (1) based on the premise $\mathbf{B}$, $C$ entails all the positive examples, denoted by $\forall \rho: \mathbf{B}, C\models\rho$, and (2) based on the premises, $C$ does not entail any of the negative examples, denoted by $\forall \eta: \mathbf{B}, C \not\models\eta$. 

Differentiable logic machines (DLM)~\cite{zimmer2023differentiable} model is a state-of-the-art neural program induction model for the ILP tasks that realizes first-order logic (FOL) rules in a neural manner. Specifically, the DLM model uses neural networks as computation units that implement ``soft'' logic operators. Then based on the computation units, the DLM model can approximate the forward-chaining mechanism of FOL by stacking multiple layers of computation units. It takes the truth value matrices of the background predicates as input and outputs the truth value matrix of the target predicate.
With the strong inductive bias of the model design, DLM has been proven to achieve superior performance and generalization on benchmarks that demand strong reasoning ability such as relational reasoning and decision-making tasks.
\vspace{-1em}
\subsection{Large Language Models}
Large language models (LLMs) denote the pre-trained Transformer architecture-based AI models.
By leveraging large amounts of multi-modal data and the pre-training \& fine-tuning learning techniques, the LLMs are reported to achieve state-of-the-art performance on many downstream tasks (\eg~machine translation, numerical reasoning, code clone detection, \etc). Specifically, the GPT model family (\eg~GPT-3.5, GPT-4, \etc) are pre-trained with an autoregressive language modeling objective.The formal definition of the autoregressive loss $\mathcal{L}_{aut}$ is as follows:
\begin{equation}
    \mathcal{L}_{aut} = - \sum_{t=2}^{n} \log p(y_t | y_{t-1}, ..., y_1)
\end{equation}
Given a sentence of $n$ tokens, the model is trained to maximize the likelihood of the ground-truth token $y_t$ of the current time step $t$ based on its previous sequence $y_{t-1}, ..., y_1$. 
In-context few-shot (zero-shot) \emph{prompting} is a recently proposed technique that conditions the LLM with some initial input to improve performance. The input can be a few examples (few-shot)~\cite{brown2020language, wei2022chain} or instructions regarding the task (zero-shot)~\cite{wei2021finetuned}.

\section{Relational Reasoning Ability Evaluation Pipeline}
 In this section, we introduce the universal relational reasoning ability evaluation pipeline for the LLMs and neural program induction models (NPIMs). The overview of the pipeline is shown in \Cref{fig:overview}, which contains four major components: (1) sample generator, (2) (truth value (tv), natural language (nl)) modality compiler, and (3) the evaluation module. The content of this section is arranged as follows: we first illustrate the details of the relational reasoning benchmark which serves as the backend of the sample generator (\Cref{benchmark}). Secondly, we introduce the sample generator which generates random data points represented in the form of truth-value matrix prompting (\Cref{generator}). Then the details of the \emph{(tv, nl)}-modality compiler are introduced. It transforms truth value prompting into corresponding natural language prompting to evaluate the models' relational reasoning ability using different data modalities (\Cref{complier}). Finally, we illustrate the evaluation module which allows measuring models' results from multiple aspects (\Cref{evaluation}).

\subsection{Relational Reasoning Benchmark}
\label{benchmark}

There are many available relational reasoning benchmarks adopted by previous literature, \eg~family tree reasoning~\cite{dongneural,evans2018learning}, general graph reasoning~\cite{DNC,dongneural,zimmer2023differentiable}, relational queries~\cite{thakkar2021example}, \etc. In this work, we conduct experiments on two broadly used relational reasoning benchmarks that are adopted by previous state-of-the-art neural program induction models~\cite{dongneural,zimmer2023differentiable,evans2018learning,manhaeve2021neural}: family tree reasoning and general graph reasoning.
\paragraph{Family tree reasoning.} The family tree reasoning benchmark consists of tasks that require the model to induce programs that deduce more complex relations based on some basic properties of family members or relations between them. Specifically, a family tree is represented with four basic predicates: \texttt{IsMother}$(x, y)$, \texttt{IsSon}$(x, y)$, \texttt{IsSon}$(x, y)$, \texttt{IsDaughter}$(x, y)$. 
\Eg~\texttt{IsMother}$(x, y)$ is \texttt{True} if $y$ is $x$'s mother, the semantics of the other basic predicates are similar. This benchmark contains 5 target predicates to induce. The details are as follows:

\begin{itemize}
[topsep=2pt,itemsep=2pt,partopsep=0ex,parsep=0ex,leftmargin=*]
  \item \texttt{HasFather}$(x)$: the semantics of \texttt{HasFather}$(x)$ is to determine whether $x$ has a father. The ground-truth program to induce is: 
  \begin{equation}
      \texttt{ HasFather }(x) \leftarrow \exists y, \texttt{ IsFather }(x, y)
  \end{equation}
  \item \texttt{HasSister}$(x)$: the semantics of this predicate is to determine whether $x$ has a sister. The ground-truth program to induce is: 
  \begin{equation}
      \texttt{HasSister}(x) \leftarrow \exists y, z,\texttt{IsDaughter}(z, y) \wedge \texttt{IsMother}(x, z)
  \end{equation}
  \item \texttt{IsGrandparent}$(x, y)$: the semantics of this predicate is to determine whether $y$ is the  grandparent of $x$. The ground-truth program to induce is: 
  \begin{equation}
\begin{aligned}
\texttt{IsGrandparent}(x, y) & \leftarrow \exists z,((\texttt{IsSon}(y, z) \wedge \texttt{IsFather}(x, z)) \\
& \vee(\texttt{IsDaughter}(y, z) \wedge \texttt{IsMother}(x, z)))
\end{aligned}
  \end{equation}
  \item \texttt{IsUncle}$(x, y)$: the semantics of this predicate is to determine if $y$ is the uncle of $x$. The ground-truth program to induce is: 
  \begin{equation}
      \begin{aligned}
& \texttt{IsUncle}(x, y) \leftarrow \exists z,((\texttt{ IsMother }(x, z) \wedge \texttt{Invented}(z, y))) \\
& \vee(\texttt{ IsFather }(x, z) \wedge \texttt{Invented}(z, y)) \\
& \texttt{Invented}(x, y) \leftarrow \exists z,((\texttt{IsSon}(z, y) \wedge \texttt{IsSon}(z, x)) \\
& \vee(\texttt{IsSon}(z, y) \wedge \texttt{IsDaughter}(z, x))) \\
&
\end{aligned}
  \end{equation}
  \item \texttt{IsMGUncle}$(x, y)$: the semantics of this predicate is to determine whether $y$ is the maternal great uncle of $x$. The ground-truth program to induce is: 
  \begin{equation}
      \texttt{IsMGUncle}(x, y) \leftarrow \exists z,(\texttt{IsMother}(x, z) \wedge \texttt{IsUncle}(z, y))
  \end{equation}
\end{itemize}


\paragraph{General graph reasoning.} The general graph reasoning benchmark consists of tasks that require the models to infer the logic of high-level target predicates that describe properties/relations of a graph based on a basic predicate: \texttt{HasEdge}$(x, y)$ (\ie~whether there is an undirected edge between node $x$ and $y$ in the graph). This benchmark contains 4 target predicates to infer. The details are as follows:

\begin{itemize}
[topsep=2pt,itemsep=2pt,partopsep=0ex,parsep=0ex,leftmargin=*]
  \item \texttt{4-Connectivity}$(x, y)$: the semantics of \texttt{4-Connectivity}$(x, y)$ is to determine whether there exists a path between node $x$ and node $y$ within 4 edges. The ground-truth program to induce is: 
  \begin{equation}
      \begin{aligned}
& \texttt{ 4-Connectivity }(x, y) \leftarrow \exists z,(\texttt{ HasEdge }(x, y) \vee \\
& \texttt{ Invented }(x, y) \vee(\texttt{ Invented }(x, z) \wedge \texttt{HasEdge}(z, y)) \vee \\
& (\texttt{ Invented }(x, z) \wedge \texttt{ Invented }(z, y))) \\
& \texttt{Invented}(x, y) \leftarrow \exists z,(\texttt{HasEdge}(x, z) \wedge \texttt{HasEdge}(z, y)) \\
&
\end{aligned}
  \end{equation}
  \item \texttt{6-Connectivity}$(x, y)$: the semantics of \texttt{6-Connectivity}$(x, y)$ is to determine whether there exists a path between node $x$ and node $y$ within 6 edges. The ground-truth program to induce is: 
  \begin{equation}
      \begin{aligned}
& \texttt{ 6-Connectivity }(x, y) \leftarrow \exists z,(\texttt{ HasEdge }(x, y) \vee \\
& \texttt { Invented1 }(x, y) \vee \texttt{Invented2}(x, y) \vee \\
& (\texttt {Invented1}(x, z) \wedge \texttt {Invented1}(z, y)) \vee \\
& (\texttt { Invented2 }(x, z) \wedge \texttt {Invented1}(z, y)) \vee \\
& (\texttt {Invented2}(x, z) \wedge \texttt{Invented2}(z, y))) \\
& \texttt {Invented1}(x, y) \leftarrow \exists z,(\texttt{HasEdge}(x, z) \wedge \texttt{HasEdge}(z, y)) \\
& \texttt{Invented2}(x, y) \leftarrow \exists z,(\texttt{HasEdge}(x, z) \wedge \texttt{Invented1}(z, y)) \\
&
\end{aligned}
  \end{equation}
  \begin{figure}[t] 
\centering 
\includegraphics[width=0.45\textwidth]{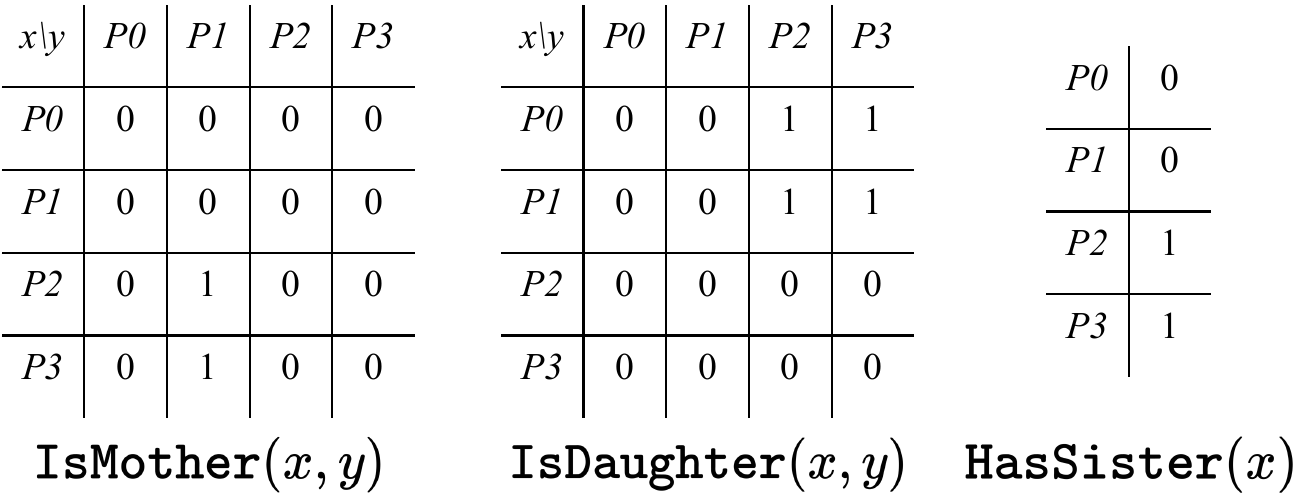} 
\caption{Truth value prompting of a \texttt{HasSister} task sample. Given the truth values of the input predicates \texttt{IsMother} and \texttt{IsDaughter}, the model is required to deduce the results of the target predicate \texttt{HasSister}.} 
\label{fig:tv_case} 
\vspace{-1.5em}
\end{figure}
  \item \texttt{1-Outdegree}$(x)$: the semantics of this predicate is to determine whether the outdegree of node $x$ in a graph is exactly 1. The ground-truth program to induce is: 
  \begin{equation}
\begin{aligned}
\texttt { 1-Outdegree }(x) \leftarrow \exists y, \forall z,(\texttt{HasEdge}(x, y) \wedge \neg \texttt{HasEdge}(x, z))
\end{aligned}
\label{eq:1-outdegree}
\end{equation}
\item {2-Outdegree}$(x)$: the semantics of this predicate is to determine whether the outdegree of node $x$ in a graph is exactly 2. The ground-truth program to induce is: 
\begin{equation}
\begin{aligned}
\texttt { 2-Outdegree }(x) & \leftarrow \exists z,w, \forall y(\neg \texttt{HasEdge}(x, y) \\
& \wedge \texttt{HasEdge}(x, z) \wedge \texttt{HasEdge}(x, w))
\end{aligned}
  \end{equation}
\end{itemize}


\vspace{-1em}
\subsection{Sample Generator}
\label{generator}
Given the specifications of each relational reasoning benchmark, the sample generator implements a truth value generator tool. Specifically, when it is called, the truth value generator tool generates the truth value matrices that represent the input \& output relations of the desired program to be induced. \Cref{fig:tv_case} shows a concrete truth value table representation of a \texttt{HasSister} task sample. It represents a family with four members $\{P0, P1, P2, P3\}$. $P0$ is the mother, $P1$ is the father, $P2$ and $P3$ are the two daughters of them. The input prompting therefore contains the truth value matrices of the input basic predicate \texttt{IsMother}$(x,y)$, \texttt{IsDaughter}$(x,y)$ \footnote{The matrices of \texttt{IsFather}$(x,y)$, \texttt{IsSon}$(x,y)$ are not presented for better illustration}. \Eg~for the \texttt{IsMother}$(x,y)$, since $P1$ is the mother of both $P2$ and $P3$, the items within the matrix that represent \texttt{IsMother}$(P2,P1)$ and \texttt{IsMother}$(P3,P1)$ equal to 1.  Then, given the input representation, the models are required to deduce the truth value matrix of the desired target predicate \texttt{HasSister}.

\vspace{-0.5em}
\subsection{(\emph{tv}, \emph{nl})-Modality Compiler}
\label{complier}
\begin{figure}[t] 
\centering
\includegraphics[width=0.39\textwidth]{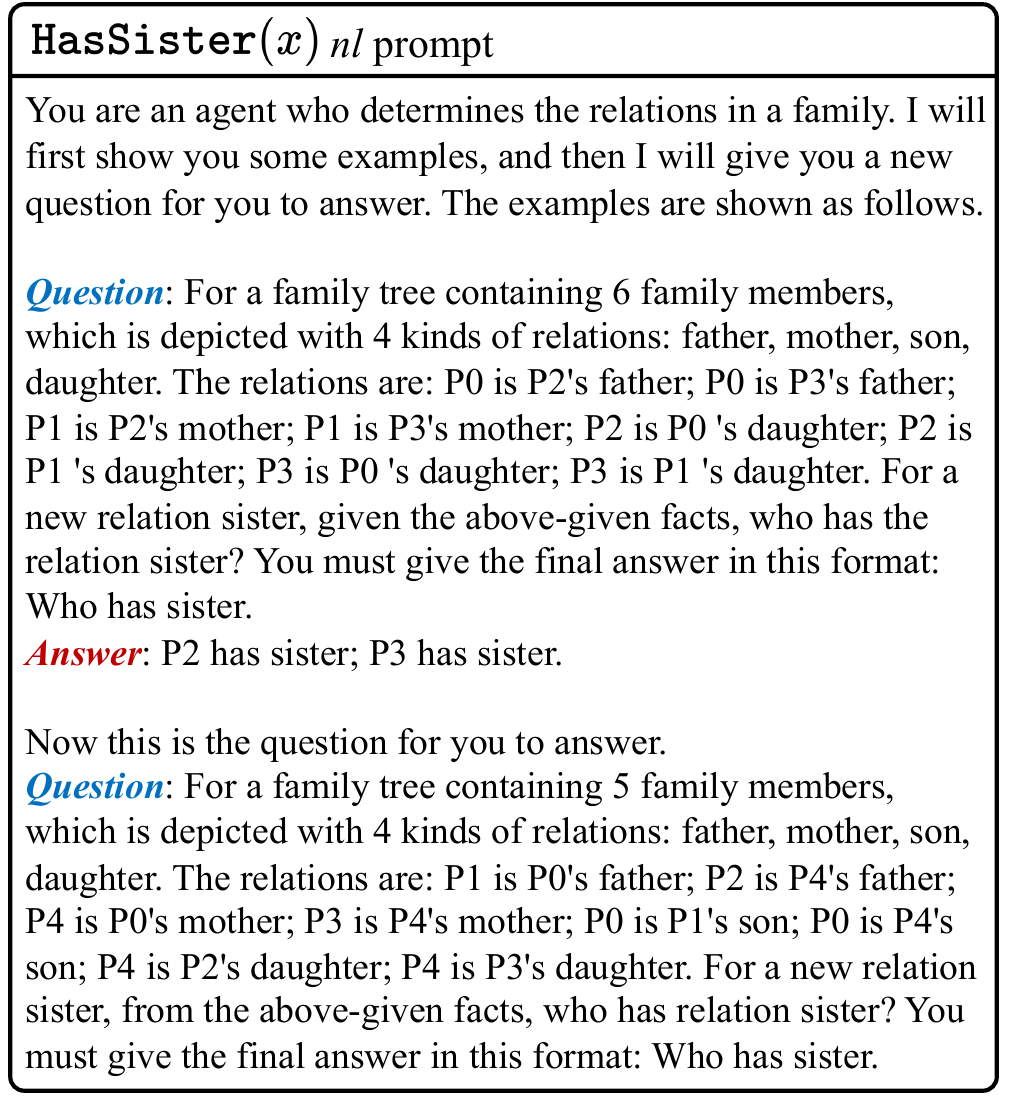}
\caption{Natural language prompting of a \texttt{HasSister} task sample.}
\label{fig:nl_case}
\vspace{-1.5em}
\end{figure}

As the LLMs are natural language models whose primary training data source is natural language, we aim to evaluate their performance when the samples are represented in the form of natural language. Therefore, we implement a (\emph{tv}, \emph{nl})-modality compiler to convert the sample from truth value prompting to natural language prompting. Specifically, the semantics of the predicates are directly conveyed via natural language prompting as background and the LLMs are then asked to answer by listing all the pairs that satisfy the target predicate. \Eg~the example in \Cref{fig:tv_case} is converted into the natural language prompting shown in \Cref{fig:nl_case}.

\vspace{-0.5em}
\subsection{Evaluation Module}
\label{evaluation}
We implement the evaluation module for evaluations of results from both the truth value source and natural language source. This module allows evaluating test performance under the independent and identically distribution (IID) setting and generalization under the out-of-distribution (OOD) setting. Specifically, for the family tree benchmark, IID is defined as the test sample whose number of family members is the same as that of the training sample. OOD is defined as the test sample with a larger family size than that of the training sample. Similarly, for the general graph reasoning benchmark, IID refers to the test sample with the same number of graph nodes as the training data while OOD refers to those with a larger number of graph nodes.
\section{Experimental Settings}
\subsection{Data setup.} We illustrate the details of the data setup of the two benchmarks used in our study. 
For all the tasks of the family tree reasoning benchmark, the number of family members for the few-shot learning (training) data is set to 10. We use data samples with the same number of family members ($n=10$) for IID performance evaluation; and samples with a larger number of family members ($n=20$) for OOD generalization evaluation. Similarly, for the general graph reasoning benchmarks, the number of nodes of each graph for the few-shot learning is set to 10. The number of nodes of each IID test sample is the same as the training samples ($n=10$), and each OOD test graph contains 20 ($n=20$) nodes.
For standard prompting, the number of training samples is set to 10, i.e., 10 families/graphs. The numbers of IID and OOD test samples are set to 10. For chain-of-thought prompting, due to the limitation of tokens, the number of training samples is set to one. The numbers of IID and OOD test samples are set to 10. As an example, for the IID test, 10 queries are conducted, each containing all the training samples and one unique testing sample. For the truth value prompting, we use Python list to represent the matrices. We conduct all experiments on a server with 48 cores of Intel(R) Xeon(R) Silver 4214 CPU @ 2.20GHz, 4 Quadro RTX 8000 GPUs, and 256G RAM.
\vspace{-1.0em}
\begin{figure}[t] 
\centering
\includegraphics[width=0.39\textwidth]{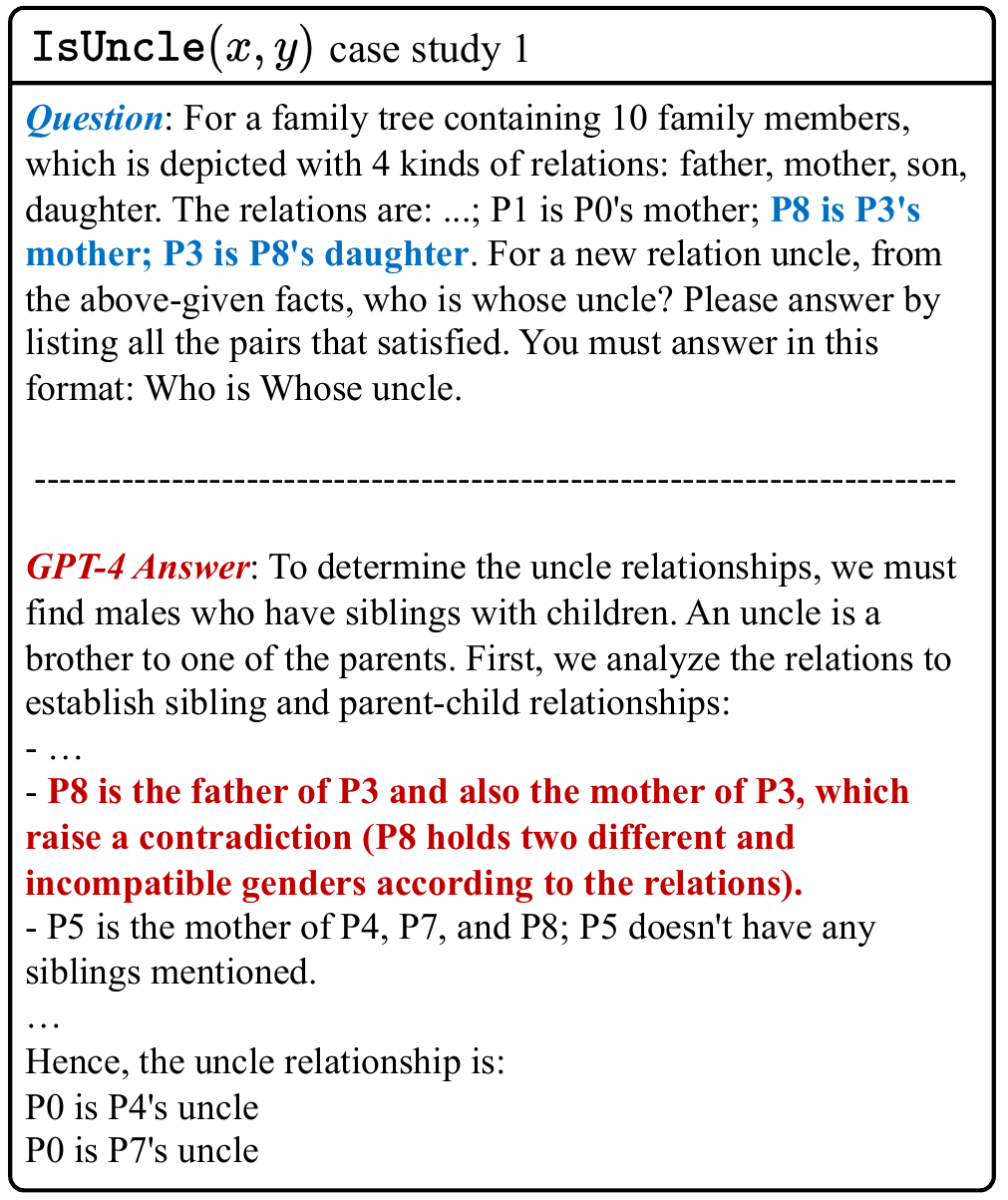}
\caption{An input-conflicting reasoning case of the GPT-4 model.}
\label{fig:casestudy}
\vspace{-1.5em}
\end{figure}
\subsection{Model setup.} 
To comprehensively evaluate the relational reasoning abilities of the LLMs and NPI models, we choose five representative LLMs baseline models and one state-of-the-art NPI model. Specifically, for LLMs, we evaluate the GPT-3.5 Turbo, GPT-4, and GPT-4 Turbo, and the open-sourced Llama 2 (7B) and Llama 2 (13B) models. For the NPI, we evaluate the differentiable logic machines (DLM) model. The details of the models are as follows:

\textbf{GPT-3.5 Turbo:} The GPT-3.5 Turbo LLM is a Transformer-based decoder network with 175 billion parameters, which is pre-trained on 45 terabytes of text data from various sources, including books, articles, and websites. It is additionally fine-tuned on specific tasks such as language translation, summarization, and question-answering and allows a context window (\ie~maximum input sequence length) of up to 16,385 tokens.

\textbf{GPT-4:} GPT-4 LLM is a successor of GPT-3.5 with an estimated 100 trillion parameters. It is a multimodal model capable of analyzing text, images, and voice data. 
It is capable of longer sequences compared to the GPT-3.5 model, which is up to 8,192 tokens.

\textbf{GPT-4 Turbo:} GPT Turbo is an enhanced version of GPT-4, featuring a significantly expanded context window (32k tokens). This enables it to efficiently process a substantial volume of data in a single run. Additionally, GPT Turbo possesses up-to-date knowledge of global events until April 2023.

\begin{table*}[!htbp]
\centering
\caption{Performance and generalization results of the LLMs using standard natural language prompting and DLM using few-shot
training. Red number denotes the best performance/generalization result among all the LLMs for a task, number in the grey box denotes the best result among all the evaluated baselines.}
  \resizebox{\textwidth}{!}{
\begin{tblr}{
  cells = {c},
  cell{1}{1} = {r=2}{},
  cell{1}{2} = {c=2}{},
  cell{1}{4} = {c=2}{},
  cell{1}{6} = {c=2}{},
  cell{1}{8} = {c=2}{},
  cell{1}{10} = {c=2}{},
  cell{1}{12} = {c=2}{},
  vline{3,5,7,9,11} = {1}{},
  vline{4,6,8,10,12} = {2-11}{},
  hline{1,12} = {-}{0.2em},
  hline{2} = {1-13}{},
  hline{3} = {-}{},
}
Task           & GPT-4 Turbo &          & GPT-4    &         & GPT3.5 Turbo  &         & llama-7b &         & llama-13b &         & DLM      &          \\
               & n=10        & n=20     & n=10     & n=20    & n=10    & n=20    & n=10     & n=20    & n=10      & n=20    & n=10     & n=20     \\
\texttt{HasFather}     & \textcolor{red}{100.00\%}    & 99.17\%  & \textcolor{red}{100.00\%} & \textcolor{red}{99.60\%} & 77.30\% & 69.52\% & 61.34\%  & 46.87\% & 47.66\%   & 59.29\% & \colorbox{lightgray}{100.00\%} & \colorbox{lightgray}{100.00\%} \\
\texttt{HasSister}     & \textcolor{red}{74.17\%}     & \textcolor{red}{81.41\%}  & 70.61\%  & 81.18\% & 46.89\% & 54.96\% & 46.77\%  & 40.49\% & 48.17\%   & 24.19\% & \colorbox{lightgray}{100.00\%} & \colorbox{lightgray}{100.00\%} \\
\texttt{IsGrandparent}   & \textcolor{red}{57.00\%}     & \textcolor{red}{35.42\%}  & 45.83\%  & 13.98\% & 32.37\% & 9.54\%  & 8.60\%   & 2.28\%  & 5.22\%    & 2.78\%  & \colorbox{lightgray}{100.00\%} & \colorbox{lightgray}{100.00\%} \\
\texttt{IsUncle}          & 26.67\%     & \textcolor{red}{17.08\%}  & \textcolor{red}{49.05\%}  & 10.15\% & 2.86\%  & 2.15\%  & 2.00\%   & 0.00\%  & 0.00\%    & 3.15\%  & \colorbox{lightgray}{85.00\%}  & \colorbox{lightgray}{29.32\%}   \\
\texttt{IsMGUncle}       & 10.00\%     & 10.00\%  & \textcolor{red}{48.33\%}  & \colorbox{lightgray}{\textcolor{red}{16.67\%}} & 0.00\%  & 3.33\%  & 0.00\%   & 0.00\%  & 0.00\%    & 0.00\%  & \colorbox{lightgray}{55.24\%}  & 11.33\%   \\
\texttt{4-Connectivity} & 62.43\%     & 11.77\%  & \textcolor{red}{73.28\%}  & \textcolor{red}{12.08\%} & 44.25\% & 6.29\%  & 28.40\%  & 9.39\%  & 61.48\%   & 26.80\% & \colorbox{lightgray}{80.82\%}  & \colorbox{lightgray}{56.58\%}  \\
\texttt{6-Connectivity} & 63.01\%     & 1.43\%   & \textcolor{red}{73.98\%}  & \textcolor{red}{37.85\%} & 25.67\% & 6.08\%  & 37.86\%  & 8.06\%  & 61.03\%   & 23.61\% & \colorbox{lightgray}{83.26\%}  & \colorbox{lightgray}{62.95\%}  \\
\texttt{1-Outdegree}    & \textcolor{red}{100.00\%}    & \textcolor{red}{100.00\%} & 57.77\%  & 64.47\% & 3.33\%  & 4.00\%  & 24.99\%  & 8.89\%  & 47.06\%   & 13.17\% & \colorbox{lightgray}{100.00\%} & \colorbox{lightgray}{100.00\%} \\
\texttt{2-Outdegree}    & \textcolor{red}{88.57\%}     & \colorbox{lightgray}{\textcolor{red}{86.59\%}}  & 86.67\%  & 75.74\% & 6.67\%  & 0.00\%  & 31.00\%  & 0.00\%  & 34.94\%   & 15.34\% & \colorbox{lightgray}{100.00\%} & 70.73\%  
\end{tblr}
}
\label{tab:standard_nl}
\vspace{-1em}
\end{table*}

\textbf{Llama 2:}
Llama 2 is an open-sourced pre-trained foundation model which adopts the standard Transformer architecture, It introduces using the Grouped Query Attention (GQA) and is capable of processing sequences with longer context length (4096 tokens) compared with the Llama 1 model (2048 tokens). It comes in four model sizes: 7B, 13B, 34B, and 70B. In this work, we evaluate the 7B and 13B models.

\textbf{DLM:} 
DLM is a neural program induction model (NPIM) architecture that approximates the inductive definition of logic formulas. DLM is the current state-of-the-art NPIM which achieves great performance and generalization of tasks ranging from relational reasoning to decision making. We strictly follow all the model setups in the original DLM paper for the evaluation. 
For all the tasks, the DLM model is trained on 10 training samples (the same set of samples used for LLMs' in-context few-shot prompting) for at most 500 epochs, and we early stop the training if the training loss is lower than $10^{-8}$ following the original DLM paper~\cite{zimmer2023differentiable}. 
We use the F1-score as the evaluation metric for all experiments since the number of positive and negative samples is imbalanced.

\begin{figure}[t] 
\centering
\includegraphics[width=0.39\textwidth]{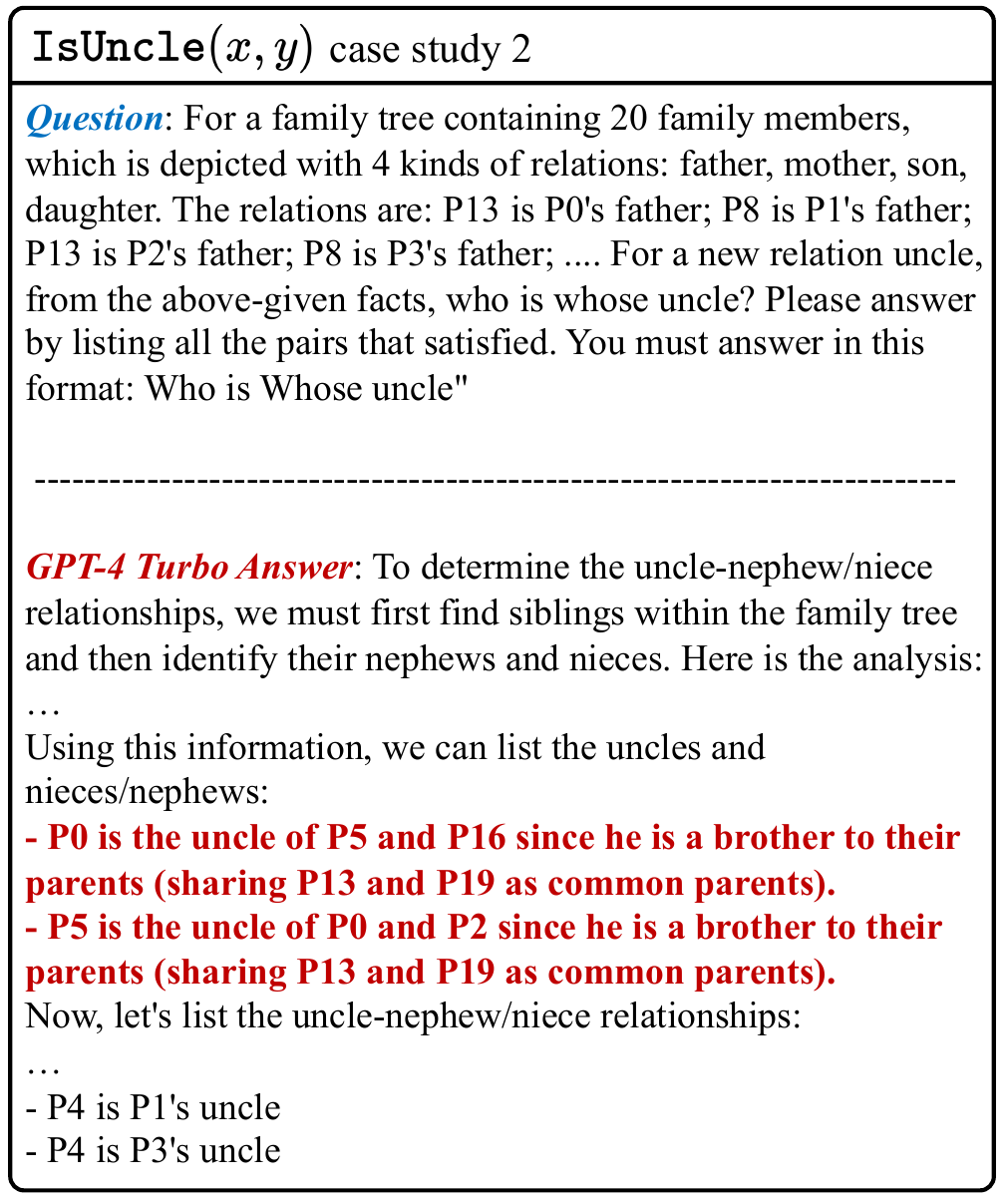}
\caption{A common sense-conflicting reasoning case of the GPT-4 Turbo model.}
\label{fig:casestudy2}
\vspace{-1.5em}
\end{figure}

\section{EVALUATION RESULTS}
In this section, we study the following research questions:
\begin{itemize}[topsep=2pt,itemsep=2pt,partopsep=0ex,parsep=0ex,leftmargin=*]
  \item \textbf{RQ1:} How good is LLMs' relational reasoning ability with standard natural language prompting?
  \item \textbf{RQ2:} How good is LLMs' relational reasoning ability with truth value prompting?
  \item \textbf{RQ3:} Is the state-of-the-art \emph{prompting} technique effective in boosting LLMs' relational reasoning ability?
\end{itemize}

\vspace{-1em}
\subsection{RQ1: How good is LLMs' relational reasoning ability with standard natural language prompting?}

\Cref{tab:standard_nl} shows the in-context few-shot prompting results of the LLMs using standard natural language prompting and the few-shot training results of the DLM. Overall, we have the following findings. 

Among all the evaluated LLMs models, GPT-4 Turbo and GPT-4 perform the best in terms of both test performance and OOD generalization.
Their superiority can be attributed to the fact that the two models are much larger in size and have access to a much larger amount of pretraining data than other LLMs baselines. Furthermore, we observe that although the LLMs can achieve great performance/generalization on relatively easier tasks, their results compromise significantly on tasks that require more complex task-solving logic (\ie~tasks that contain more predicates in the ground-truth program). For example, while the GPT-4 model achieves 100\% F1-score in terms of performance on the \texttt{HasFather} task, it only achieves 49.05\% on the \texttt{IsUncle}. To better understand why the LLMs perform poorly on harder tasks, we conduct two case studies of the GPT-4 and GPT-4 Turbo models on the \texttt{IsUncle} test sample, shown in \Cref{fig:casestudy} and \Cref{fig:casestudy2}. 
For \Cref{fig:casestudy}, though it is stated in the input prompt that ``P8 is P3's mother; P3 is P8's daughter'' (bold blue text), GPT-4 generates an input-conflicting reasoning process stating that ``P8 is the father of P3 and also the mother of P3'' (bold red text). 
For \Cref{fig:casestudy2}, as shown in GPT-4 Turbo's answer, the model generates a common sense-conflicting reasoning process that claims that P0 and P5 are each other's uncle (bold red text).
These hallucination phenomena would therefore undermine the model's relational reasoning ability, leading to erroneous conclusions.


Finally, though the DLM model is much smaller in size compared to the LLMs (DLM has only 60k parameters, while the smallest evaluated LLM has 7 billion parameters (llama-7B)), it manages to achieve the best results with only few-shot training. Note that the hardware requirement required for training the DLM model is much lower than even loading the LLMs (\eg~ it only requires 429 MiB of GPU memory for training DLM while 36,957 MiB is required for loading llama-7B). The reason is that the logical reasoning knowledge injected in the DLM's model design allows it to conduct robust program induction which makes it suitable for the relational reasoning tasks.

\begin{tcolorbox}[size=title,opacityfill=0.1,breakable]
\textbf{Answer to RQ1}: 
GPT-4 and GPT-4 Turbo models present the strongest relational reasoning ability among all the evaluated LLMs. However, the LLMs' performance/generalization drops significantly on tasks that require complex task-solving logic. Besides, the relational reasoning ability of the LLMs' in-context few-shot prompting is generally worse than the few-shot trained DLM model which is much less hardware-demanding.
\end{tcolorbox}

\begin{table}
\caption{Performance and generalization results of DLM and LLMs using truth table prompting.}
\centering
 \resizebox{0.5\textwidth}{!}{
\begin{tblr}{
  cells = {c},
  cell{1}{1} = {r=2}{},
  cell{1}{2} = {c=2}{},
  cell{1}{4} = {c=2}{},
  cell{1}{6} = {c=2}{},
  vline{3,5} = {1}{},
  vline{4,6} = {2-11}{},
  hline{1,12} = {-}{0.2em},
  hline{2} = {2-7}{},
  hline{3} = {-}{},
}
Task           & GPT-4-Turbo &         & GPT-3.5 Turbo &      & DLM      &          \\
               & n=10        & n=20    & n=10    & n=20 & n=10     & n=20     \\
\texttt{HasFather}     & 79.17\%   & 76.99\% & 47.57\% & N/A  & \colorbox{lightgray}{100.00\%} & \colorbox{lightgray}{100.00\%} \\
\texttt{HasSister}     & 47.06\%   & 50.49\% & 38.41\% & N/A  & \colorbox{lightgray}{100.00\%} & \colorbox{lightgray}{100.00\%} \\
\texttt{IsGrandparent}   & 9.52\%    & N/A     & 6.62\%  & N/A  & \colorbox{lightgray}{100.00\%} & \colorbox{lightgray}{100.00\%} \\
\texttt{IsUncle}          & 6.67\%    & N/A     & 0.00\%  & N/A  & \colorbox{lightgray}{85.00\%}  & \colorbox{lightgray}{29.32\%}  \\
\texttt{IsMGUncle}        & 0.00\%    & N/A     & 0.00\%  & N/A  & \colorbox{lightgray}{55.24\%}  & \colorbox{lightgray}{11.33\%}  \\
\texttt{4-Connectivity} & 60.76\%   & N/A     & 53.35\% & N/A  & \colorbox{lightgray}{80.82\%}  & \colorbox{lightgray}{56.58\%}  \\
\texttt{6-Connectivity} & 65.77\%    & N/A     & 51.67\% & N/A  & \colorbox{lightgray}{83.26\%}  & \colorbox{lightgray}{62.95\%}  \\
\texttt{1-Outdegree}    & 98.90\%     & 99.19\% & 50.05\% & N/A  & \colorbox{lightgray}{100.00\%} & \colorbox{lightgray}{100.00\%} \\
\texttt{2-Outdegree}    & \colorbox{lightgray}{100.00\%}    & \colorbox{lightgray}{95.31\%} & 42.65\% & N/A  & 100.00\% & 70.73\%  
\end{tblr}
}
\label{tab:tv}
\vspace{-1.3em}
\end{table}

\begin{table*}[!htbp]
\caption{Performance and generalization results of DLM and LLMs using chain-of-thought prompting.}
\centering
  \resizebox{\textwidth}{!}{
\begin{tblr}{
  cells = {c},
  cell{1}{1} = {r=2}{},
  cell{1}{2} = {c=2}{},
  cell{1}{4} = {c=2}{},
  cell{1}{6} = {c=2}{},
  cell{1}{8} = {c=2}{},
  cell{1}{10} = {c=2}{},
  cell{1}{12} = {c=2}{},
  vline{3,5,7,9,11} = {1}{},
  vline{4,6,8,10,12} = {2-11}{},
  hline{1,12} = {-}{0.2em},
  hline{2} = {1-13}{},
  hline{3} = {-}{},
}
Task           & GPT-4-Turbo &         & GPT-4    &          & GPT-3.5 Turbo &         & llama-7b &         & llama-13b &         & DLM      &          \\
               & n=10        & n=20    & n=10     & n=20     & n=10    & n=20    & n=10     & n=20    & n=10      & n=20    & n=10     & n=20     \\
\texttt{HasFather}     & 100.00\%    & 99.57\%  & \textcolor{red}{100.00\%} & \textcolor{red}{100.00\%}  & 76.32\% & 85.04\%  & 59.49\%  & 63.77\%  & 86.98\%    & 64.30\%  & \colorbox{lightgray}{100.00\%} & \colorbox{lightgray}{100.00\%} \\
\texttt{HasSister}     & \textcolor{red}{86.39\%}     & \textcolor{red}{76.26\%} & 79.88\%   & 73.60\%  & 16.39\% & 20.26\% & 57.94\%   & 47.98\%  & 52.91\%   & 53.33\%  & \colorbox{lightgray}{100.00\%} & \colorbox{lightgray}{100.00\%} \\
\texttt{IsGrandparent}   & 63.57\%     & 38.38\%  & \textcolor{red}{70.32\%}  & \textcolor{red}{39.02\%}  & 11.96\% & 0.82\%  & 2.99\%   & 1.03\%  & 7.08\%    & 2.75\%  & \colorbox{lightgray}{100.00\%} & \colorbox{lightgray}{100.00\%} \\
\texttt{IsUncle}          & 44.44\%     & \textcolor{red}{23.84\%}  & \textcolor{red}{96.67\%}   & 16.35\%   & 23.33\%  & 1.55\%  & 0.00\%   & 0.00\%  & 0.00\%    & 1.43\%  & \colorbox{lightgray}{85.00\%}  & \colorbox{lightgray}{29.32\%}   \\
\texttt{IsMGUncle}        & 10.00\%      & \textcolor{red}{10.00\%}  & \textcolor{red}{40.00\%}   & 4.68\%   & 0.00\%  & 0.00\%  & 2.00\%   & 0.00\%  & 0.00\%    & 0.00\%  & \colorbox{lightgray}{55.24\%}  & \colorbox{lightgray}{11.33\%}   \\
\texttt{4-Connectivity} & 26.00\%     & 3.44\%  & \textcolor{red}{79.78\%}   & \textcolor{red}{19.33\%}  & 1.18\%  & 28.41\%  & 40.82\%  & 18.64\%  & 39.33\%   & 27.13\%  & \colorbox{lightgray}{80.82\%}  & \colorbox{lightgray}{56.58\%}  \\
\texttt{6-Connectivity} & 54.82\%     & 3.50\%  & \textcolor{red}{86.31\%}  & \textcolor{red}{11.46\%}  & 2.61\%  & 24.27\%  & 50.88\%   & 18.98\%  & 33.55\%   & 23.02\% & \colorbox{lightgray}{83.26\%}  & \colorbox{lightgray}{62.95\%}  \\
\texttt{1-Outdegree}    & \textcolor{red}{86.39\%}     & \textcolor{red}{83.18\%} & 38.19\%  & 38.45\%  & 5.00\%   & 2.86\%  & 24.81\%  & 20.51\%  & 14.00\%   & 6.19\%  & \colorbox{lightgray}{100.00\%} & \colorbox{lightgray}{100.00\%} \\
\texttt{2-Outdegree}    & \colorbox{lightgray}{\textcolor{red}{100.00\%}}    & \colorbox{lightgray}{\textcolor{red}{89.24\%}}  & 76.67\%  & 19.16\%  & 5.00\%  & 0.00\%  & 28.71\%  & 0.00\%  & 20.67\%   & 4.72\%  & 100.00\% & 70.73\%  
\end{tblr}
}
\label{tab:cot}
\vspace{-1em}
\end{table*}

\subsection{RQ2: How good is LLMs' relational reasoning ability with truth value prompting?}
We further assess the relational reasoning ability of the LLMs when using the same data modality as the DLM model (\ie~representing input/output predicate relations with truth value matrices). \Cref{tab:tv} shows the in-context few-shot learning results of the LLMs using truth value prompting. The result is marked as N/A if the model cannot generate the complete output array. We omit the results of the GPT-4, llama-7b, and llama-13b models as all their results are N/A. The reason for the complete N/A results for these three models is because they are relatively small in terms of context window whereas the GPT-4 Turbo and GPT-3.5 Turbo are capable of processing long context window and generating long output sequence\footnote{Note that for all the evaluated tasks, the ground-truth output truth value matrices required to generate never exceed the maximum output token length of any LLMs.}.

Interestingly, we observe that the results of GPT-4 Turbo and GPT-3.5 Turbo on the family tree reasoning tasks decrease as compared to the corresponding standard natural language prompting results. This is because the natural language pretraining data regarding family relations is common and sufficient which allows the LLMs to perform well with natural language prompting on the family tree reasoning tasks. While the truth value pretraining data is much fewer comparatively which makes it challenging for the LLMs to reason well with truth value prompting. Whereas interestingly, the IID performance is consistent or even improves on the general graph reasoning tasks as compared to the standard natural language prompting results, the GPT-4 Turbo even manages to achieve the best performance and generalization on the \texttt{2-outdegree} task. We believe that LLMs have the ability to perform graph reasoning with truth value matrices, which makes them promising for truth value-related tasks (\eg~logic synthesis~\cite{hassoun2001logic}). We leave the further investigation to future work.



\begin{tcolorbox}[size=title,opacityfill=0.1,breakable]
\textbf{Answer to RQ2}: 
LLMs with small context window are incapable of conducting relational reasoning with truth value
prompting, while the LLMs with large context window present consistent or even improved IID performance on the general graph reasoning tasks compared to the standard natural language prompting results.
\end{tcolorbox}

\begin{figure}[t] 
\centering
\includegraphics[width=0.39\textwidth]{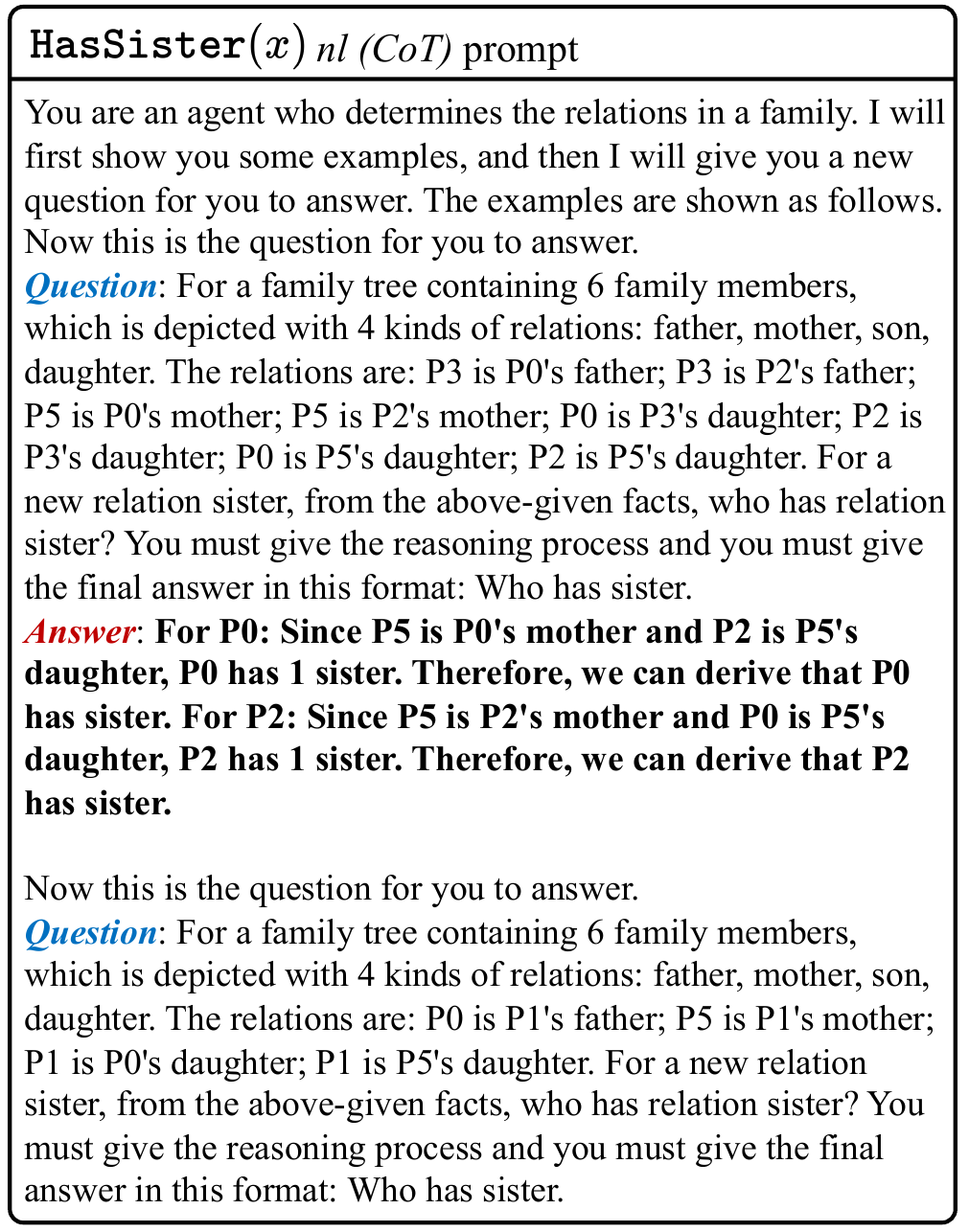}
\caption{Chain-of-thought prompting of a \texttt{HasSister} sample.}
\label{fig:nl_case_cot}
\vspace{-1.5em}
\end{figure}

\subsection{RQ3: Is the state-of-the-art \emph{prompting} technique effective in boosting LLMs' relational reasoning ability?}

We further evaluate whether the recently proposed \emph{chain-of-thought} (CoT)~\cite{wei2022chain} prompting can improve the LLMs' relational reasoning ability. Concretely, each exemplar in few-shot prompting is augmented with the direct cause-effect logic that represents the deduction process of the answer. \Cref{fig:nl_case_cot} shows an example CoT prompting of a \texttt{HasSister} sample, the bold text represents the example deduction process. The evaluation results of LLMs using natural language CoT prompting are shown in \Cref{tab:cot}. 
Though it is reported in previous literature that CoT can generally improve the LLMs' performance on benchmarks such as arithmetic, commonsense, \etc. We observe that for the evaluated relational reasoning benchmarks, CoT cannot consistently improve the performance/generalization of the LLMs. For some tasks, the results are even significantly decreased, \eg~the performance and generalization of GPT-4 Turbo on the \texttt{1-outdegree} task are decreased by 13.61\%, 16.82\% respectively.

\begin{tcolorbox}[size=title,opacityfill=0.1,breakable]
\textbf{Answer to RQ3}: 
The CoT prompting technique is not generally effective for the relational reasoning benchmarks. The CoT prompting can even significantly decrease the LLMs' performance/generalization for some tasks.
\end{tcolorbox}



\section{Related Works}


\subsection{LLMs for logical reasoning}
With the rapid advancement of LLMs, there has been a recent surge in research that leverages LLMs to solve logical reasoning tasks, such as common-sense reasoning~\cite{talmor2018commonsenseqa,sakaguchi2021winogrande}, arithmetic reasoning~\cite{cobbe2021arithmetic,patel2021arithmetic}, symbolic reasoning~\cite{srivastava2022beyond}, \etc.
It is observed in previous works that the standard prompting (only questions are given) is not sufficient as the model performs poorly~\cite{wei2022chain}. Aiming to improve the LLMs' performance, Brown \etal~\cite{brown2020language} propose prompting the model with a few
input–output exemplars demonstrating the task, which is proven to be effective for a range of
simple question-answering tasks. Wei \etal~\cite{wei2022chain} propose providing the model with the concrete chain-of-thought prompting (\ie I/O examples with the corresponding explicit reasoning steps) to the LLMs. It is demonstrated that the chain-of-thought prompting technique improves performance on arithmetic reasoning, commonsense reasoning, and symbolic reasoning tasks. To improve LLMs' zero-shot performance, Kojima \etal~\cite{kojima2022large} further introduces zero-shot-CoT prompting which is able to improve performance by simply enhancing the input prompt with a sentence ``Let's think step by step''. Though the LLMs manage to achieve great accomplishments on these tasks, Valmeekam \etal~\cite{valmeekam2023planbench} find that the LLMs still perform poorly on simple decision-making tasks such as the \emph{blocksworld} which requires model arrange a set of given blocks in a stack in a particular order. To the best of our knowledge, our work is the first study that evaluates LLMs' relational reasoning ability.

\subsection{Neural Program Induction \& Synthesis} 
Leveraging neural networks for program induction and program synthesis has become increasingly popular. Recent work has demonstrated the effectiveness of neural network-based program induction and synthesis methods in relational reasoning tasks with better noise tolerance and less manual effort~\cite{devlin2017robustfill,evans2018learning,glanois2022neuro}. Dong \etal~\cite{dongneural} propose Neural Logic Machines (NLMs) which approximate logic predicates and logic rules with neural modules and achieves state-of-the-art performance on the program induction tasks. To improve the interpretability of the NLM model, Zimmer \etal~\cite{zimmer2023differentiable} propose using soft logic operators as the computation unit which allows the model to generate interpretable logic programs. Evans \etal~\cite{evans2018learning} propose a differentiable implementation of inductive logic programming ($\partial$ILP), which is capable of synthesizing fully interpretable programs with high noise tolerance to noisy input examples. Jiang \etal~\cite{jiang2019neural} introduce using $\partial$ILP to synthesize white-box logic programs as reinforcement learning agent's policies. Cao \etal~\cite{cao2022galois} propose a sketch-based hierarchical program synthesis framework for complex sequential decision-making problems based on $\partial$ILP, which achieves high performance and generalizability.

\section{Conclusion}
In this work, we propose a general and extensible relational reasoning ability assessment pipeline for the LLMs and the neural program induction models. Based on the pipeline, we conduct the first comprehensive evaluations of the LLMs' relational reasoning ability. Our results show that when using standard natural language prompting, the relational reasoning ability of LLMs' in-context few-shot prompting is generally far from satisfaction compared with the program induction model which is much smaller in size. While the LLMs with large context window present consistent or even improved IID performance on the general graph reasoning tasks. And we further show that the current state-of-the-art prompting technique is not generally effective for improving LLMs' relational reasoning ability. 

\section{Acknowledgement}
This research is supported by the National Research Foundation, Singapore, and the Cyber Security Agency under its National Cybersecurity R\&D Programme (NCRP25-P04-TAICeN) and NRF Investigatorship NRF-NRFI06-2020-0001, and the RIE2020 Industry Alignment Fund – Industry Collaboration Projects (IAF-ICP) Funding Initiative, as well as cash and in-kind contributions from the industry partner(s). Any opinions, findings and conclusions or recommendations expressed in this material are those of the author(s) and do not reflect the views of National Research Foundation, Singapore and Cyber Security Agency of Singapore.


\bibliographystyle{ACM-Reference-Format}
\bibliography{main}

\end{document}